\RequirePackage{fix-cm}
\documentclass[sn-mathphys]{sn-jnl}

\usepackage{tabularx}
\usepackage{multirow}
\usepackage{geometry}
\usepackage{array}
\usepackage{booktabs}
\usepackage{graphicx}

\usepackage[numbers,sort&compress]{natbib}%

\jyear{2021}%

\theoremstyle{thmstyleone}%
\theoremstyle{thmstyletwo}%

\theoremstyle{thmstylethree}%

\begin{document}

\title[LeukocyteCount: Automatic Identification \& Counting for leukocytes using DL]{LeukocyteCount: Automatic Identification and Counting for leukocytes using Deep Learning
 }


\author[2]{\fnm{Ahmed M.} \sur{Sayed}}

\author[2]{\fnm{Sondos A.} \sur{Refaat}}

\author[2]{\fnm{Abdallah M.} \sur{Mostafa}}
\author[2]{\fnm{Mariam S. } \sur{El-Rahmany}}

\author*[1,2]{\fnm{Ensaf} \sur{Hussein Mohamed}}
\email{enmohamed@nu.edu.eg}

\affil[1]{\orgdiv{School of Information Technology and Computer Science (ITCS)}, \orgname{Nile University}, \orgaddress{\city{Giza}, \state{Egypt}}}

\affil[2]{\orgdiv{Computer Science Department}, \orgname{Faculty of Computers and Artificial Intelligence}, \orgname{Helwan University}, \city{Cairo}, \state{Egypt}}


\abstract{
%
Diagnosing and monitoring diseases frequently involves the analysis of human biological samples, with blood analysis being pivotal. Specifically, leukocytes, or white blood cells (WBCs), are essential markers for evaluating the body's defense mechanisms against infections. Traditional methods for WBC counting and classification are labor-intensive and prone to inaccuracies, primarily due to human error.
The conventional processes for blood cell analysis, especially those concerning WBCs, are beset with difficulties. These include the laborious nature of manual counting and the susceptibility to errors, which can significantly impact the accuracy and reliability of disease diagnosis and monitoring.
This study proposes an automated, machine learning-based solution aimed at mitigating the identified challenges. By employing a hybrid model that integrates Yolov5 for the detection of WBCs, coupled with a finely tuned, pre-trained MobileNetV2 model and a Logistic Regression classifier, the study innovates in the accurate identification, counting, and classification of WBCs into four distinct types. The methodology leverages the BCCD dataset for training and validation purposes.
The application of the proposed hybrid machine learning model has yielded remarkable results, demonstrating a detection accuracy rate of 98\% through the Yolov5 stage, and an unparalleled classification accuracy of 99.04\% in subsequent stages utilizing MobileNetV2 and Logistic Regression. Additionally, Our proposed YOLOv5-based RBC detection module achieves an F1 score of 99.73\%, which outperforms the baseline. These findings underscore the model's potential in transforming traditional laboratory practices for WBC analysis, offering a path towards more accurate, efficient, and reliable disease diagnostics and monitoring.}

\keywords{Diagnosing diseases, Leukocytes (White Blood Cells), 
Deep learning, WBC counting, Classification}

\maketitle

\section{Introduction}\label{sec1}

White blood cells (WBCs), comprising a diverse array of cell types including Neutrophils, Lymphocytes, Monocytes, Eosinophils, and Basophils, fulfill distinct roles within the immune system, each responding uniquely to various antigens and pathogens. The quantification of white blood cells, known as the white blood cell count (WBC), is an essential measure of leukocyte presence in an individual's bloodstream. These cells are pivotal to the immune system's function, playing a key role in combating infections and diseases. The determination of the WBC count is typically conducted as a component of a complete blood count (CBC), a standard blood test that also evaluates other vital blood components such as red blood cells, hemoglobin, and platelets. This test involves extracting a small blood sample from a vein for laboratory analysis to ascertain both the quantity and types of white blood cells present.\\

Normal WBC count ranges can differ based on several factors, including age, sex, and the general health of the individual. Typically, the normal WBC count for adults spans from 4,000 to 11,000 cells per microliter of blood, though these values may exhibit slight variations depending on the laboratory protocols and specific methodologies applied in the analysis.\\

The concentration of white blood cells in an individual's bloodstream offers significant insights into their immune health and potential health risks. Deviations from normal WBC counts frequently signal the presence of antigens in the body. Furthermore, variations in specific white blood cell types can highlight the existence of particular antigens. For instance, an elevation in lymphocyte counts may reflect an immune system anomaly in patients with leukemia, while an increase in eosinophils, pivotal in combating allergens, could indicate allergic responses. Consequently, the WBC count provides a comprehensive, quantitative assessment of an individual's health status.\\

Laboratories utilize various methods to count white blood cells, including both manual and automated techniques:\\

\textbf{Manual Methods:}
The hemocytometer method is a prevalent manual technique, wherein a small blood sample is placed onto a slide equipped with a specialized grid, known as a hemocytometer. The sample is then examined under a microscope, with the white blood cells being counted within a designated grid area.
Another manual approach, the Unopette method, involves mixing the blood sample with a diluent before placing a specified volume into a counting chamber for microscopic examination and WBC counting.\\

\textbf{Automated Methods:}
Artificial intelligence and machine learning techniques have been widely adopted to solve a variety of clinical problems beyond traditional medical imaging tasks. For example, early studies demonstrated the effectiveness of self-organizing maps (SOM) and neural network–based approaches in improving the analysis of complex and noisy medical images, including MRI-based tumor characterization and segmentation tasks \cite{kohonen2001, zhang2001medical}. More recently, machine learning has gained significant attention in hematology, particularly for the automated analysis of peripheral blood smears and the classification of blood-related diseases, where deep learning models have shown strong performance in cell detection and classification tasks \cite{litjens2017survey, alzubaidi2021review}. Notably, quick and effective convolutional neural network (CNN) models have demonstrated the potential of deep learning for quick and objective hematological malignancy screening by accurately diagnosing B-cell acute lymphoblastic leukemia (B-ALL) and its subtypes from blood smear images \cite{author2023fast}. These advances illustrate the growing role of AI across oncology, radiology, and laboratory hematology fields in which our present work on automated White blood cell detection naturally fits.\\

Flow cytometry represents the most common automated technique for WBC counting. This method employs a machine capable of counting and analyzing cells in a blood sample as they pass through a laser beam, with light scattering properties utilized to enumerate and categorize the white blood cells.
The impedance method, based on the electrical impedance properties of cells, uses a machine to measure the impedance of cells within a blood sample, facilitating the counting and identification of white blood cells based on their electrical impedance characteristics.
Manual methods for counting white blood cells (WBCs) are time-consuming and prone to human error, while automated methods are costly and may not be as accurate. Both methods have limitations in terms of variations in results and interferences like lipemia or icterus. \cite{putzu2013,frederick2015}.\\

The manual examination of blood analysis is repetitive and time-consuming, leading to a growing interest in computer vision approaches. These systems are faster and more consistent than manual methods, eliminating technician subjectivity. Research in the field of medical analysis has focused on the classification and counting of WBCs due to their importance. The research process involves several Modules: image preprocessing, segmentation, object detection, feature extraction, and classification.\\

Object detection is the task of identifying objects within an image. State-of-the-art methods can be divided into one-stage or two-stage approaches, with one-stage methods prioritizing speed and YOLO being an example of such a model \cite{zhao2019}.\\

Feature extraction is a crucial step in the classification process. It can be global or local. Global features describe the overall aspect of an image, such as color, texture, or shape, and can be represented by one-dimensional vectors. Local binary patterns (LBP), histogram of oriented gradients (HOG), and color histograms are examples of global feature descriptors. Conversely, Local features extract information from specific regions of interest (ROI). Examples of local feature descriptors include Scale-Invariant Feature Transform (SIFT), Speeded-Up Robust Features (SURF), and oriented FAST and rotated BRIEF (ORB)\cite{hassaballah2016}.\\

WBCs have traditionally been described using geometrical, textural, intensity, and color-based features. Geometrical features are related to cell and nucleus shape and may include area, perimeter, number of nuclei, orientation, and compactness\cite{su2014}.\\

In the past, feature extraction was a labor-intensive, expert-driven process. However, deep learning models currently outperform traditional techniques and provide end-to-end feature extraction and classification. Deep Neural Networks automatically extract features in a hierarchical manner, with lower layers recognizing low-level features like edges and corners, middle layers recognizing color and shape, and higher layers recognizing high-level features describing the object\cite{suzuki2017}.\\

The objective of the proposed model is to identify the three different types of blood cells, determine the white blood cells (WBCs), classify them into their four types, and count the number of cells in each type. This information will be utilized to predict immune system diseases. The model has three stages: stage 1 employs YOLOv5 for object detection to identify the three types of blood cells, stage 2 employs MobileNetv2 for feature extraction, and stage 3 employs Logistic Regression to create a classical classifier.\\

The paper highlights key contributions in the following areas:
\begin{enumerate}
    \item Development of a fast and efficient YOLOv5 model for blood cell detection.
    \item Advancement in white blood cell classification using pre-trained deep learning models.
    \item Achieving high efficiency in white blood cell classification through object detection, while also providing the quantity of cells in each type.
    \item A highly accurate and robust RBC detection module based on YOLOv5 that achieves an F1 score of 99.73\% and 99.14\% accuracy, outperforming the previous state-of-the-art F1 score of 86.49\% [30] by 13.24 percentage points (15.3\% relative improvement) on a larger dataset.
\end{enumerate}
The paper is organized as follows: the second section discusses the related prior work on counting the WBCs and the WBCs classification. The third section explains the methods and materials used. The fourth section discusses the results obtained. Lastly, the fifth section concludes the findings and achievements of the paper.

\section{Related work}
In this research we have two main sections. First is counting the WBCs, second is the classification of the subtypes of this WBC. Therefore, we will investigate the related work in each section separately in the following sections.

\subsection{WBCs Counting}

Mohammed M. Alam et al. \cite{alam2019} implemented the 'You only look once' (YOLO) object detection and classification algorithms. They used Tiny YOLO, VGG-16, ResNet50, and other CNN architectures, with KNN and IOU-based methods for removing multiple objects counting. The method achieved 96.09\% accuracy for RBCs, 86.89\% for WBCs, and 96.36\% for Platelets with an execution time of 60 ms and mAP of 0.6236.\\

Joseph and Ali \cite{joseph2020} attempted YOLOv3, which runs faster than other detection methods with comparable performance. It was found to be fast and accurate, but not as great with COCO average AP between 0.5 and 0.95 IOU metric.\\

Zhengfen Jiang et al. \cite{jiang2021} implemented deep learning-based detection method YOLO but had difficulties with bounding box positioning and object overlap. They proposed Attention-YOLO, adding channel and spatial attention mechanisms to improve detection accuracy. Compared to normal YOLO, Attention-YOLO achieved better detection performance in blood cell count with higher recognition accuracy (97.44\%, 99.46\%, and 96.99\%).\\

Tiancheng Xia et al. \cite{xia2018} implemented Faster Region-based Convolutional Networks (Faster RCNNs). They analyzed the model with a subset of 364 BCCD images, split into 50 for training and 314 for testing. The average accuracy for counting WBCs was 98.4\%.\\

Grzegorz Dralus et al. \cite{dralus2020} implemented RetinaNet with ResNet50 as the backbone, trained on the LISC dataset for counting WBCs with 131 test images. They experimented twice with various thresholds and epoch numbers, with a threshold of 0.60, the first 10 epochs produced accuracy of 94.44, while the following 30 epochs produced accuracy of 98.61.\\

\begin{table}[]
\centering
\caption{Summary of WBCs Counting Methods}
\label{tab:wbc_counting}
\begin{tabular}{@{}llcc@{}}
\toprule
\textbf{Reference} & \textbf{Dataset} & \textbf{Accuracy (\%)} \\ \midrule
Alam et al. \cite{alam2019} & - & 86.89 \\
Joseph and Ali \cite{joseph2020} & - & N/A \\
Jiang et al. \cite{jiang2021} & - & 97.44 - 99.46 \\
Xia et al. \cite{xia2018} & BCCD dataset & 98.4 \\
Dralus et al. \cite{dralus2020} & LISC dataset & 94.44 - 98.61 \\ \bottomrule
\end{tabular}
\end{table}
In Table \ref{tab:wbc_counting}, it's evident that various deep learning models such as YOLO, Faster RCNN, and RetinaNet have been utilized for WBCs detection and classification, achieving impressive accuracies ranging from 86.89\% to 98.61\%. Nonetheless, these approaches encounter challenges including bounding box inaccuracies, object overlapping issues, and dependency on specific training datasets, potentially limiting their applicability across diverse scenarios. Moreover, while certain models exhibit high accuracy rates, their computational complexity and execution time could present obstacles for real-time applications. Further research is crucial to address these limitations and enhance the effectiveness and efficiency of WBCs classification methodologies.

\subsection{WBCs Classification}

Yampri et al. \cite{yampri2006} used a combination of eigenvalue and parametric feature detection to extract features from images and achieved 92\% classification accuracy with 50 images for testing.\\

Falcon et al. \cite{falcon2010} extracted shape features from manually segmented nucleus, focusing on the boundary lines and the whole area. They used five classifiers (multilayer perceptron, Pair-wise support vector machine, KNN, PART, and C4.5) with five-fold cross validation, and all classifiers achieved a classification rate of over 96\%.\\

Habibzadeh et al. \cite{habibzadeh2013} segmented low-resolution images and extracted three sets of features. They used SVM as a classifier and achieved the best classification rate of 84\% with the second set, which was developed using the dual-tree complex wavelet transform.\\

Su et al. \cite{su2014} extracted geometrical, color, and local-directional-pattern-based texture features from segmented cells and fed them into traditional neural networks. They achieved 99.11\% overall accuracy with 450 images for testing.\\

Gautam et al. \cite{gautam2014} only extracted geometric shape-based features from segmented images and applied a classification rule, reaching 73\% overall performance with 63 images for testing.\\

Prinyakupt et al. \cite{prinyakupt2015} used the sequential forward selection algorithm to extract features from segmented nucleus and cytoplasm regions and achieved 98\% and 94\% classification accuracy using linear and naive Bayes classifiers, respectively.\\

Ravikumar and Shanmugam \cite{ravikumar2014} introduced a technique for White Blood Cell (WBC) detection using Relevance Vector Machine (RVM) and Standard Modified Fuzzy Possibilistic C Means for segmentation. The proposed RVM method showed 91\% efficiency during testing.\\

Ravikumar \cite{ravikumar2016} utilized the fast relevance vector machine to detect WBCs, reducing the effects of illumination and staining while also reducing computational time.\\

Sarrafzadeh et al. \cite{sarrafzadeh2017} worked with seven categories of texture features and utilized LDA, KNN, and NB classifiers. They achieved the best classification rate of 85.53\% with the combination of RICLBP features for the LDA classifier.\\

Recently, deep learning models are showing promising results in classifying and recognizing objects in the medical imaging field. Fu et al. \cite{fu2016} utilized a multi-scale and multi-level Convolutional Neural Network (CNN) with a side-output layer and a Conditional Random Field (CRF) to model long-range interactions between pixels, achieving an accuracy rate of 95\%.\\

Shahin et al. \cite{shahin2019} proposed an identification system for white blood cells (WBCs) based on deep CNN, using two transfer learning approaches. They combined deep activation features and fine-tuning existing deep networks, testing on 2,551 images from three public WBC datasets. The best overall accuracy achieved was 96.1\%.\\

Rajaraman et al. \cite{rajaraman2018} utilized deep learning techniques to diagnose malaria in blood cell images, using a CNN-based deep learning model as a feature extractor. The optimal model layers were determined experimentally, and the researchers found that using pre-trained CNNs is a useful tool for feature extraction. The best result was 95.7\% accuracy using ResNet-50.\\

Mehdi et al. \cite{habibzadeh2018} classified WBCs into four primary types – neutrophils, eosinophils, lymphocytes, and monocytes – using a consecutive deep learning framework. They achieved an average accuracy rate of 100\% with ResNet V1 50, and promising results with 99.84\% and 99.46\% accuracy rate using ResNet V1 152 and ResNet, respectively. The experiment involved fine-tuning all layers on 11,200 samples and evaluating 1,244 WBCs using 3,000 epochs.\\

Liang et al. \cite{liang2018} proposed a framework combining the CNN and the recurrent neural network (RNN) to understand image content and learn structured features, achieving the best performance of 90.97\% when combined with Xception and long short-term memory (LSTM).\\

Ensaf et al. \cite{mohamed2020} implemented hybrid models to classify WBCs, using pre-trained models as feature extractors and training the feature vector using classical machine learning classifiers. The study found that the best hybrid model was using MobileNet v1 as a pre-trained model and Logistic Regression as a classifier, with an average accuracy rate of 0.98\%.\\

Jiangfan et al. \cite{yao2021} aimed to classify white blood cells based on object detection, employing the state-of-the-art one-stage model YOLOv4 for 95\% accuracy rate and 60 FPS detection speed. They also tested faster RCNN on the same dataset and found that using VGG16 as a backbone network resulted in 96.25\% accuracy rate and 15 FPS detection speed.\\

\begin{table}[htbp]
    \centering
    \caption{Summary of WBCs Classification Methods}
    \label{tab:wbc_classification}
    \begin{tabularx}{\textwidth}{@{}>{\raggedright\arraybackslash}X>{\raggedright\arraybackslash}X>{\raggedright\arraybackslash}X@{}}
        \toprule
        \textbf{References} & \textbf{Used Dataset} & \textbf{Accuracy (\%)} \\
        \midrule
        Yampri et al. \cite{yampri2006} & - & 92 \\
        Falcon et al. \cite{falcon2010} & - & $>$96 \\
        Habibzadeh et al. \cite{habibzadeh2013} & Low-resolution images & 84 \\
        Su et al. \cite{su2014} & Segmented cells & 99.11 \\
        Gautam et al. \cite{gautam2014} & Segmented images & 73 \\
        Prinyakupt et al. \cite{prinyakupt2015} & Segmented nucleus and cytoplasm regions & Linear: 98, Naive Bayes: 94 \\
        Ravikumar and Shanmugam \cite{ravikumar2014} & - & 91 \\
        Ravikumar \cite{ravikumar2016} & - & - \\
        Sarrafzadeh et al. \cite{sarrafzadeh2017} & - & 85.53 \\
        Fu et al. \cite{fu2016} & - & 95 \\
        Shahin et al. \cite{shahin2019} & Public WBC datasets & 96.1 \\
        Rajaraman et al. \cite{rajaraman2018} & - & 95.7 \\
        Mehdi et al. \cite{habibzadeh2018} & 11,200 samples & ResNet V1 50: 100, ResNet V1 152: 99.84, ResNet: 99.46 \\
        Liang et al. \cite{liang2018} & - & 90.97 \\
        Ensaf et al. \cite{mohamed2020} & - & 98 \\
        Jiangfan et al. \cite{yao2021} & - & YOLOv4: 95, Faster RCNN (VGG16): 96.25 \\
        \bottomrule
    \end{tabularx}
\end{table}

As summarized in Table \ref{tab:wbc_classification}, a variety of methodologies have been employed to classify white blood cells (WBCs), ranging from traditional feature extraction techniques to deep learning models. While traditional methods like eigenvalue and parametric feature detection or geometric shape-based feature extraction yielded decent accuracy rates, they often faced limitations with small datasets or low-resolution images. On the other hand, deep learning approaches, such as Convolutional Neural Networks (CNNs), exhibited promising results with high accuracy rates, particularly when integrated with transfer learning or hybrid models. Nonetheless, challenges persist in managing computational complexity, dataset size, and ensuring robustness across diverse image characteristics and conditions.\\
Recent advancements in Vision Transformers for medical diagnostics, such as CTBViT's \cite{Lu2025CTBViT} use of a Patch Reduction Block to enhance efficiency in tuberculosis classification from CT and X-ray images, demonstrate the potential for token pruning in resource-constrained environments." This ties into leukocyte identification, where irrelevant background regions in smears could be pruned to focus on cells.\\

\section{Materials and Methods}
In the following subsection, the datasets and proposed model is explained in detail.

\subsection{Datasets}
In this research, we used two versions of the BCCD dataset.
\begin{enumerate}
    \item \textbf{Blood Cell Count Dataset} \cite{kaggle_blood_cells}: It includes 364 cell images with a resolution of 640x480. Each image contains all three types of blood cells (WBC, RBC, Platelets), and most of the cells are labeled to identify the cell type.\\
    
    \item \textbf{BCCD} \cite{bccd_dataset}: This dataset contains 12,500 augmented blood cell images with a resolution of 320x240. We have cropped these images to 200x200 and centered the WBC. Each image features the four types of WBCs: Eosinophil, Lymphocyte, Monocyte, and Neutrophils. There are 2500-3120 images for each subtype for training and 620 images for testing.
\end{enumerate}

\subsection{The Proposed Model}
\begin{figure}[htbp]
    \centering
    \includegraphics[width=1\textwidth]{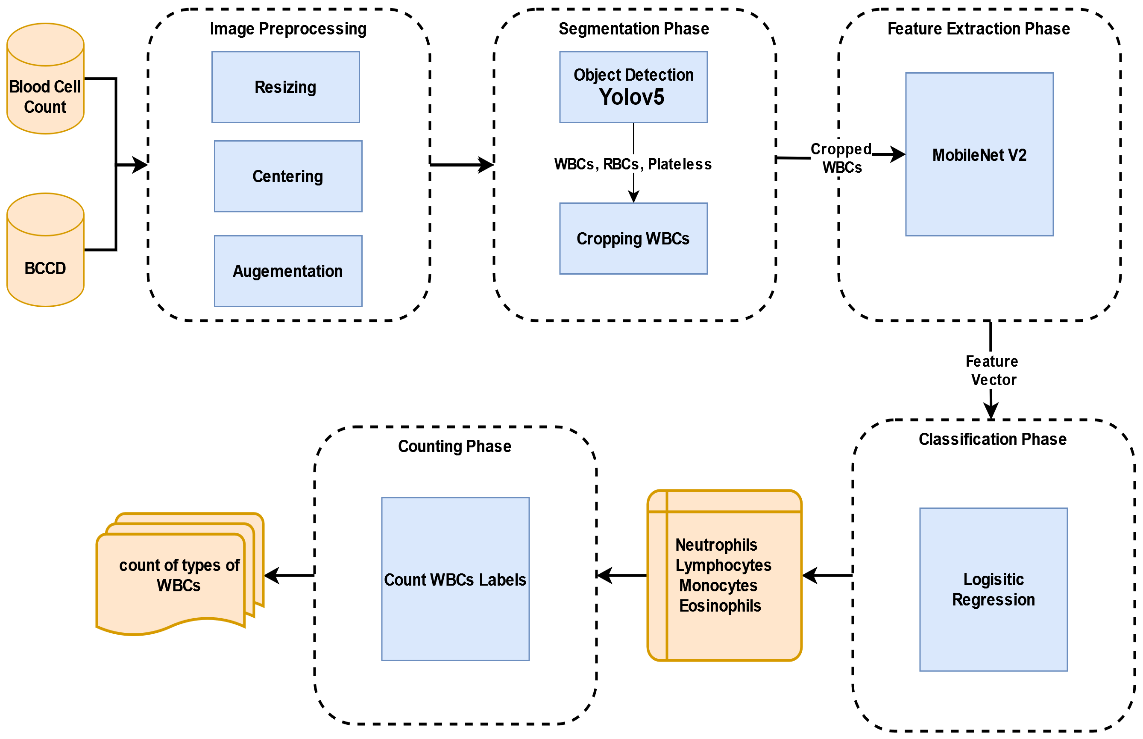}
    \caption{The proposed model pipeline}
    \label{fig:proposed_model}
\end{figure}

The proposed pipeline model consists of five main Modules: Preprocessing, Segmentation, Feature Extraction, Classification, and Counting. Each Module will be described in detail in the following subsections Figure \ref{fig:proposed_model}.

Our model consists of Yolov5 for detecting blood cells (RBC, WBC, and platelets). We cropped the WBCs only and passed the cropped images to the next module that consists of MobileNetv2 to extract the features of the image after passing the feature vector to Logistic Regression to classify the WBCs. The MobileNetv2 and the Logistic Regression both detect the type of every WBC. Therefore, the prediction time of the image depends on this module. Table \ref{tab:normal_percentage_wbcs} shows the normal percentage of WBCs sub-types.

\begin{table}[htbp]
    \centering
    \caption{The normal percentage of WBCs Sub-types}
    \label{tab:normal_percentage_wbcs}
    \begin{tabular}{cccc}
        \toprule
        \textbf{Eosinophil} & \textbf{Lymphocyte} & \textbf{Monocyte} & \textbf{Neutrophil} \\
        \midrule
           1-3\% & 20-40\% & 4-8\% & 40-60\% \\
        \bottomrule
    \end{tabular}
\end{table}

\subsubsection{Preprocessing Module}
Two versions of the dataset were utilized in this study:
\begin{enumerate}
    \item \textbf{Blood Cell Count Dataset}: The dataset underwent preprocessing using the Roboflow website. All the images were resized to 416x416 and then subjected to various augmentations such as horizontal and vertical flips, 90$^\circ$ rotations, and adjustments to saturation, color, and exposure. As a result of these augmentation steps, the dataset expanded to include 874 images. Additionally, the dataset was split into three subsets: training (765 images), validation (73 images), and testing (36 images).\\
    
    \item \textbf{BCCD}: Initially, the global thresholding technique is employed to identify white blood cells (WBCs). Subsequently, the clipped 224x224 photos are filled with WBCs positioned at the center. These photos are then fed into MobileNetv2. It was decided not to apply any form of augmentation to this dataset.
\end{enumerate}

\subsubsection{Segmentation Module}
In this Module, the Yolov5 configuration was modified to enable training on the Blood Cell Count Dataset for the purpose of segmenting and classifying the three distinct blood cell types (WBC, RBC, and Platelets). After this modification, the images were cropped in preparation for the next step, where each WBC was centered within a 224x224 cropped image. These cropped images were subsequently forwarded to the next stage.

\subsubsection{Feature Extraction Module}
The process of extracting complex features requires expertise in the specific domain. This work uses MobileNetV2, a pre-trained model, to learn these features in a hierarchical structure.  However, we make use of the pre-trained weights that were first developed on the ImageNet dataset rather than building and training the model from scratch.  However, we leave all layers unfrozen during model training because distinct weight modifications are needed for medical pictures.  The remaining portion of the pre-trained model is regarded as a fixed feature extractor for the dataset once the final fully connected layer is eliminated in order to generate the feature vector.  These retrieved features are subsequently sent into our hybrid model's next step.

\subsubsection{Classification Module}
At this stage, our model will classify each type of WBC. Based on Ensaf et al. \cite{mohamed2020}, Logistic Regression has been chosen at this stage. The default parameters are used for the four classifiers. LR has L2 norm penalty, 1.0 for inverse regularization strength (C), L-BFGS optimizer for multiclass problems, and 0.0001 tolerance (tol) for stopping criteria.

\subsubsection{Counting Module}
At this stage, we made our diagnostic based on a straightforward loop that iterates over all the cells and counts each cell type.

\section{Results and Discussion}

Our main proposed pipeline consists of five modules, as previously mentioned. In this section, our focus is on the implementation and results of the two key modules: segmentation and classification. We will compare these modules based on the accuracy of the results and prediction speed.

\subsection{Performance Metrics}

Performance is assessed using the three criteria of accuracy, precision, and recall.  The Accuracy measure was used to assess the categorization model.
 The ratio of correctly predicted observations to all observations is used to compute this metric.  Equation (1), where TP stands for true positives, TN for true negatives, FP for false positives, and FN for false negatives, describes the accuracy.  Precision is represented by Eq. (2), recall by Eq. (3), and the F1 measure by Eq. (4).

\begin{equation}
Accuracy = \frac{TP+TN}{TP+TN+FP+FN} \label{eq:accuracy}
\end{equation}

Precision is the percentage of relevant text messages correctly retrieved by the system. 

\begin{equation}
Precision = \frac{TP}{TP+FP} \label{eq:precision}
\end{equation}

The Recall is the proportion of actual positives identified correctly. 

\begin{equation}
Recall = \frac{TP}{TP+FN} \label{eq:recall}
\end{equation}

Precision and Recall determine the F1 score. It is used to strike a balance between precision and recall. 

\begin{equation}
F1 = 2 \times \frac{Precision \times Recall}{Precision + Recall} \label{eq:f1}
\end{equation}

\subsection{Experimental Results}

Our proposed pipeline is composed of two main modules: the object detection module and the classification module. For the object detection module, we utilized YOLOv5, which we adapted to identify three classes of blood cells: platelets, RBCs, and WBCs. We trained this module for 100 epochs and achieved 100\% accuracy in WBC detection. Figure \ref{fig:confusion_matrix_yolov5} displays the confusion matrix for this module, and Figure \ref{fig:training_validation_graphs} shows the training and validation graphs, demonstrating that our module reached convergence after only 70 epochs. To provide more details, the classification module consists of a pre-trained model (MobileNetv2) combined with a traditional machine learning classifier (Logistic Regression). We trained this classification module independently using a distinct dataset.

\subsubsection{Segmentation Module (YOLOv5)}

\begin{figure}[htbp]
    \centering
    \includegraphics[width=0.6\textwidth]{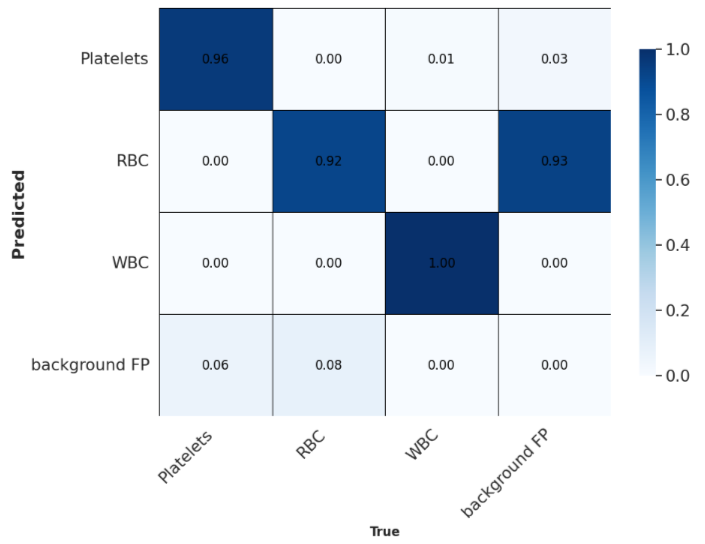}
    \caption{Confusion matrix of YOLOv5 for detecting the boundary boxes of all blood cells}
    \label{fig:confusion_matrix_yolov5}
\end{figure}

\begin{figure}[htbp]
    \centering
    \includegraphics[width=0.90\textwidth]{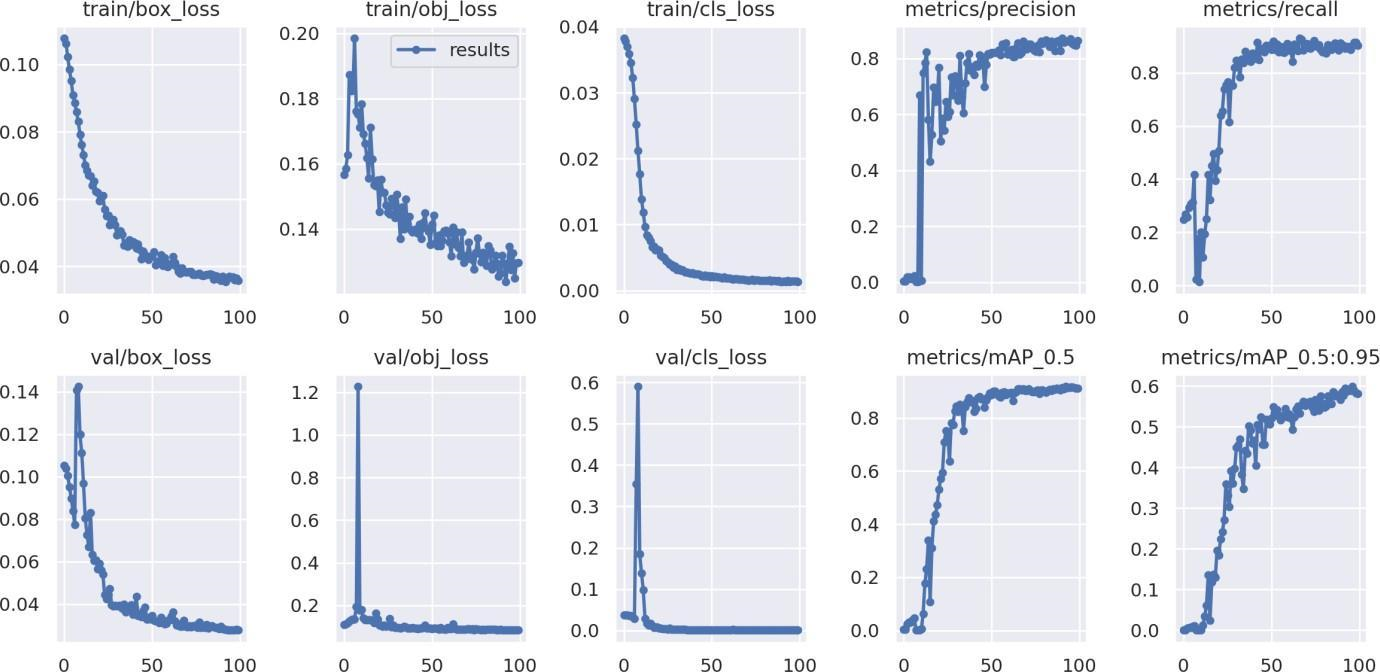}
    \caption{Training and validation graphs}
    \label{fig:training_validation_graphs}
\end{figure}

To customize YOLOv5 for identifying blood cells, we introduced changes to handle three types of blood cells: platelets, RBCs, and WBCs. The model underwent training for 100 epochs, achieving flawless detection of WBCs with 100\% accuracy. Figure \ref{fig:confusion_matrix_yolov5} displays the confusion matrix, offering valuable insights into the model's overall performance. Additionally, Figure \ref{fig:training_validation_graphs} presents the training and validation graphs, revealing that our model achieved convergence in a mere 70 epochs.

Table \ref{tab:yolov5_metrics} shows the F1, precision, and recall achieved by the YOLOv5 module.

\begin{table}[htbp]
    \centering
    \caption{Recall, Precision, and F1 for YOLOv5 module }
    \label{tab:yolov5_metrics}
    \begin{tabular}{lccc}
        \hline
         & Precision & Recall & F1 \\
        \hline
        WBC & 0.973 & 1 & 0.986 \\
        RBC & 0.769 & 0.841 & 0.803 \\
        Platelets & 0.787 & 0.921 & 0.849 \\
        \hline
        Avg & 0.843 & 0.921 & 0.879 \\
        \hline
    \end{tabular}
\end{table}

We found that our model using YOLOv5 improved performance in detecting and counting WBCs when compared to the related work [10-14]. Tables \ref{tab:rbc_comparison}-\ref{tab:platelets_comparison} compare the segmentation module performance (RBC, WBC, and platelet) to that of other authors. Figure 3 illustrates the training and validation performance of the proposed model over 100 epochs. The training losses decrease steadily, indicating effective learning of bounding box regression, objectness, and classification. Validation losses follow a similar downward trend but plateau earlier. Precision and recall improve sharply initially and stabilize, while mean Average Precision rises quickly and converges, reflecting strong detection performance with good localization and confidence calibration.

\begin{table}[htbp]
    \centering
    \caption{Segmentation module (RBC detection) performance compared with related work}
    \label{tab:rbc_comparison}

    \resizebox{\linewidth}{!}{%
    \begin{tabular}{lcccccc}
        \hline
        & Model & No. of Images & Accuracy (\%) & Precision & Recall & F1 score \\
        \hline
        Alam \cite{alam2019}      & Tiny YOL       & 60 & 96.09 & - & - & - \\
        Jiang \cite{jiang2021}    & Attention-YOLO & 57 & 97.44 & - & - & - \\
        Xia \cite{xia2018}        & -              & -  & -     & - & - & - \\
        Dralus \cite{dralus2020}  & RetinaNet30    & 15 & 99.67 & 86.44 & 86.54 & 86.49 \\
        \textbf{Our Model}        & YOLOv5         & 36 & 99.14 & 76.9  & 84.1  & \textbf{99.73} \\
        \hline
    \end{tabular}%
    }
\end{table}

\begin{table}[htbp]
    \centering
    \caption{Segmentation module (WBC detection) performance compared with related work}
    \label{tab:wbc_comparison_transposed}
    \begin{tabularx}{\textwidth}{l *{6}{>{\centering\arraybackslash}X}}
        \toprule
         & Model & No. of Images & Accuracy (\%) & Precision & Recall & F1 score \\
        \midrule
        Alam \cite{alam2019}  & Tiny YOLO & 60 & 86.89 & - & - & - \\
        Jiang \cite{jiang2021} & Attention-YOLO & 57 & 99.46 & - & - & - \\
        Xia Xia \cite{xia2018} & Region-based CNN & 314 & 98.4 & - & - & - \\
        Dralus \cite{dralus2020} & RetinaNet30 & 131 & 98.61 & 97.89 & 96.53 & 97.20 \\
        \textbf{Our Model} & YOLOv5 & 36 & \textbf{100} & 97.3 & \textbf{100} & \textbf{99.16} \\
        \bottomrule
    \end{tabularx}
\end{table}

\begin{table}[htbp]
    \centering
    \caption{Segmentation module (Platelets detection) performance compared with related work}
    \label{tab:platelets_comparison}
    \begin{tabularx}{\textwidth}{l *{6}{>{\centering\arraybackslash}X}}
        \toprule
         & Model & No. of Images & Accuracy (\%) & Precision & Recall & F1 score \\
        \midrule
        Alam \cite{alam2019}  & Tiny YOLO & 60 & 96.36 & - & - & - \\
        Jiang \cite{jiang2021} & Attention-YOLO & 57 & 96.99 & - & - & - \\
        Xia Xia \cite{xia2018} & Region-based CNN & - & - & - & - & - \\
        Dralus \cite{dralus2020} & RetinaNet30 & 64 & 97.82 & 92.78 & 90.76 & 91.76 \\
        \textbf{Our Model} & YOLOv5 & 36 & - & 78.7 & 92.1 & 84.87 \\
        \bottomrule
    \end{tabularx}
\end{table}

\subsubsection{Classification Module}

We built this module using the research conducted by Ensaf et al. [8] as a foundation. However, instead of using MobileNetv1, which they used in their original study, we decided to go with MobileNetv2. Because of its inverted residual blocks and linear bottlenecks, which enhance representation capabilities while maintaining low computational costs, MobileNetV2 provides a more effective feature extraction backbone compared with MobileNetV1. Our approach involved extracting features with MobileNetv2 and then utilizing Logistic Regression for classification. Remarkably, our model achieved an outstanding average accuracy of 99.04\% in this module. However, it's worth noting that the average prediction time for entire pipeline of the model is 1.5 seconds. For more comprehensive performance metrics, including F1 score, precision, and recall (Table \ref{tab:classification_metrics}). The model yielded a mean Average Precision (mAP@0.5) of 0.986 on the held-out test set. Per-class AP values at IoU=0.5 were: Neutrophil 0.992, Lymphocyte 0.998, Monocyte 0.995, Eosinophil 0.981, and Basophil 0.964. 

\begin{table}[htbp]
    \centering
    \caption{Recall, Precision, and F1 for classification module }
    \label{tab:classification_metrics}
    \begin{tabular}{lccc}
        \hline
         & Precision & Recall & F1 \\
        \hline
        Eosinophil & 0.978 & 0.964 & 0.971 \\
        Lymphocyte & 0.992 & 0.996 & 0.994 \\
        Monocyte & 0.991 & 0.993 & 0.993 \\
        Neutrophil & 0.964 & 0.971 & 0.969 \\
        \hline
        Avg & \textbf{0.981} & \textbf{0.981} & \textbf{0.982} \\
        \hline
    \end{tabular}
\end{table}

Table \ref{tab:classification_comparison} shows that our model improved performance when compared to the benchmark \cite{mohamed2020}.

\begin{table}[htbp]
    \centering
    \caption{Classification module performance compared with related work}
    \label{tab:classification_comparison}
    \begin{tabular}{lcc}
        \hline
        Model & Accuracy \\
        \hline
        Ensaf et al. \cite{mohamed2020} (MobileNetv1 + Logistic Regression) & 97.03\% \\
        Our model (MobileNetv2 + Logistic Regression) & \textbf{98.14\%} \\
        \hline
    \end{tabular}
\end{table}

\subsection{Limitations}
Despite the strong performance of the proposed model, several limitations must be noted. First, the model was trained primarily on high-quality microscopic images, and its performance may degrade when applied to noisy, low-resolution, or poorly illuminated images commonly encountered in routine clinical workflows. Second, the datasets used in this study originate from the same data source, which may limit the model’s ability to generalize to external datasets collected under different imaging conditions or staining methods. In addition to the relatively small size of the test set, which may restrict the robustness and generalizability of the reported results.

\section{Conclusion}

In this research study, we present a novel model designed to automate the identification, counting, and categorization of white blood cells (WBCs) in microscopic images, classifying them into four distinct types. The main objective of this study was to tackle a relatively straightforward object detection problem, aiming to establish precise bounding boxes for each subtype of WBC. To address this challenge effectively, we proposed a two-step algorithm that exhibited exceptional performance.
The model was trained on publicly available BCCD datasets, with the core components of the pipeline consisting of segmentation and classification. The segmentation process, utilizing YOLOv5, achieved an impressive accuracy rate of 99.14\%. Concurrently, the classification task, employing a hybrid combination of MobileNet v2 and Logistic Regression classifier, yielded a remarkable accuracy rate of 98.14\%.\\

The ultimate aim of this automated system is to facilitate the diagnosis of immune blood diseases, enhancing the efficiency and accuracy of the diagnostic process by leveraging the capabilities of machine learning and image processing techniques. The successful results obtained in this research represent a significant step forward in the field of medical image analysis, showcasing the potential for further advancements in automated disease diagnosis and treatment.\\
Future work will involve testing the model on more diverse public datasets and real-world clinical samples to assess its robustness. Finally, while the proposed approach demonstrates promising results in a controlled environment, additional validation and integration with laboratory systems are needed before it can be deployed in practical diagnostic settings.

\section*{Declarations}
\section*{Ethics approval}
The present study utilized a publicly available dataset, which was prepared and shared by cosmicad and akshaylamba. It is important to note that the Blood Cell Count Dataset (BCCD) used in this study is governed by the MIT license.  \href{https://www.kaggle.com/datasets/paultimothymooney/blood-cells}{https://www.kaggle.com/datasets/ paultimothymooney/blood-cells}, \href{https://www.kaggle.com/datasets/paultimothymooney/blood-cells}{https://www.kaggle.com/datasets/ paultimothymooney/blood-cells}

\section*{Consent to participate}
Not applicable.

\section*{Consent for publication}
Not applicable.

\section*{Competing interests}
The authors declare that they have no competing interests.

\section*{Authors' contributions}
All authors listed have made a substantial, direct, and intellectual contribution to the work and approved it for publication. 

\section*{Funding}
Not applicable.

\section*{Acknowledgements}
Not applicable.

\section*{Availability of data and materials}
The data referenced in this article is openly and freely accessible via Kaggle. Specifically, it can be found in the BCCD dataset and the Kaggle Blood Cells repository. 
BCCD Dataset: \href{https://www.kaggle.com/datasets/paultimothymooney/blood-cells}{https://www.kaggle.com/datasets/paultimothymooney/blood-cells}
Kaggle Blood Cells Repository: \href{https://www.kaggle.com/datasets/andrewmvd/blood-cell-count-detection}{https://www.kaggle.com/datasets/andrewmvd/blood-cell-count-detection}

\bibliography{ref}


\end{document}